\documentclass[letterpaper, 10 pt, conference]{ieeeconf}  

\IEEEoverridecommandlockouts                              

\usepackage{bm}
\usepackage{url}
\usepackage{graphicx}
\usepackage{lipsum}
\usepackage{pifont}
\usepackage{amsmath}
\usepackage{amssymb}
\usepackage{nccmath}
\usepackage{comment}
\usepackage{balance}
\usepackage{multirow}
\usepackage{textcomp}
\usepackage{multirow}
\usepackage{subcaption}
\usepackage{booktabs}
\usepackage{algorithm}
\usepackage{dblfloatfix}
\usepackage[table]{xcolor}
\usepackage[noend]{algpseudocode}

\title{\LARGE \bf
General, Single-shot, Target-less, and Automatic \\
LiDAR-Camera Extrinsic Calibration Toolbox
}

\author{Kenji Koide$^{1}$, Shuji Oishi$^{1}$, Masashi Yokozuka$^{1}$, and Atsuhiko Banno$^{1}$
\thanks{*This work was supported in part by a project commissioned by the New Energy and Industrial Technology Development Organization (NEDO).}
\thanks{$^{1}$All the authors are with the Department of Information Technology and Human Factors, the National Institute of Advanced Industrial Science and Technology, Tsukuba, Ibaraki, Japan, {\tt\small k.koide@aist.go.jp}}%
}

\begin{document}

\maketitle
\thispagestyle{empty}
\pagestyle{empty}

\definecolor{verylightgray}{gray}{0.95}
\newcolumntype{g}{>{\columncolor{verylightgray}}c}

\newcommand{\argmax}{\mathop{\rm arg~max}\limits}
\newcommand{\argmin}{\mathop{\rm arg~min}\limits}

\algnewcommand\Input{\item[\textbf{Input:}]}%
\algnewcommand\Output{\item[\textbf{Output:}]}%

\newcommand{\xmark}{\ding{55}}%

\setlength\floatsep{6pt}
\setlength\textfloatsep{6pt}

\begin{abstract}

This paper presents an open source LiDAR-camera calibration toolbox that is general to LiDAR and camera projection models, requires only one pairing of LiDAR and camera data without a calibration target, and is fully automatic. For automatic initial guess estimation, we employ the SuperGlue image matching pipeline to find 2D-3D correspondences between LiDAR and camera data and estimate the LiDAR-camera transformation via RANSAC. Given the initial guess, we refine the transformation estimate with direct LiDAR-camera registration based on the normalized information distance, a mutual information-based cross-modal distance metric. For a handy calibration process, we also present several assistance capabilities (e.g., dynamic LiDAR data integration and user interface for making 2D-3D correspondence manually). The experimental results show that the proposed toolbox enables calibration of any combination of spinning and non-repetitive scan LiDARs and pinhole and omnidirectional cameras, and shows better calibration accuracy and robustness than those of the state-of-the-art edge-alignment-based calibration method.

\end{abstract}

\section{Introduction}

LiDAR-camera extrinsic calibration is the task of estimating the transformation between the coordinate frames of a LiDAR and a camera. It is necessary for LiDAR-camera sensor fusion and is required for many applications, including autonomous vehicle localization, environmental mapping, and surrounding-object recognition.

Although LiDAR-camera calibration has been actively studied over the last decade, the robotics community still lacks a handy and complete LiDAR-camera calibration toolbox. The existing LiDAR-camera calibration frameworks require preparing a calibration target that is sometimes difficult to create \cite{Xie2022}, taking many shots of LiDAR-camera data that results in a large amount of effort \cite{Tsai2021}, or choosing a geometry-rich environment carefully \cite{Yuan2021}. Furthermore, they rarely support various LiDAR and camera projection models, such as spinning and non-repetitive scan LiDARs and ultra-wide FoV and omnidirectional cameras. We believe that the lack of an easy-to-use LiDAR-camera calibration method has been a long-standing barrier to the development of LiDAR-camera sensor fusion systems.

\begin{figure}[tb]
  \centering
  \begin{minipage}[tb]{\linewidth}
  \centering
  \includegraphics[width=\linewidth]{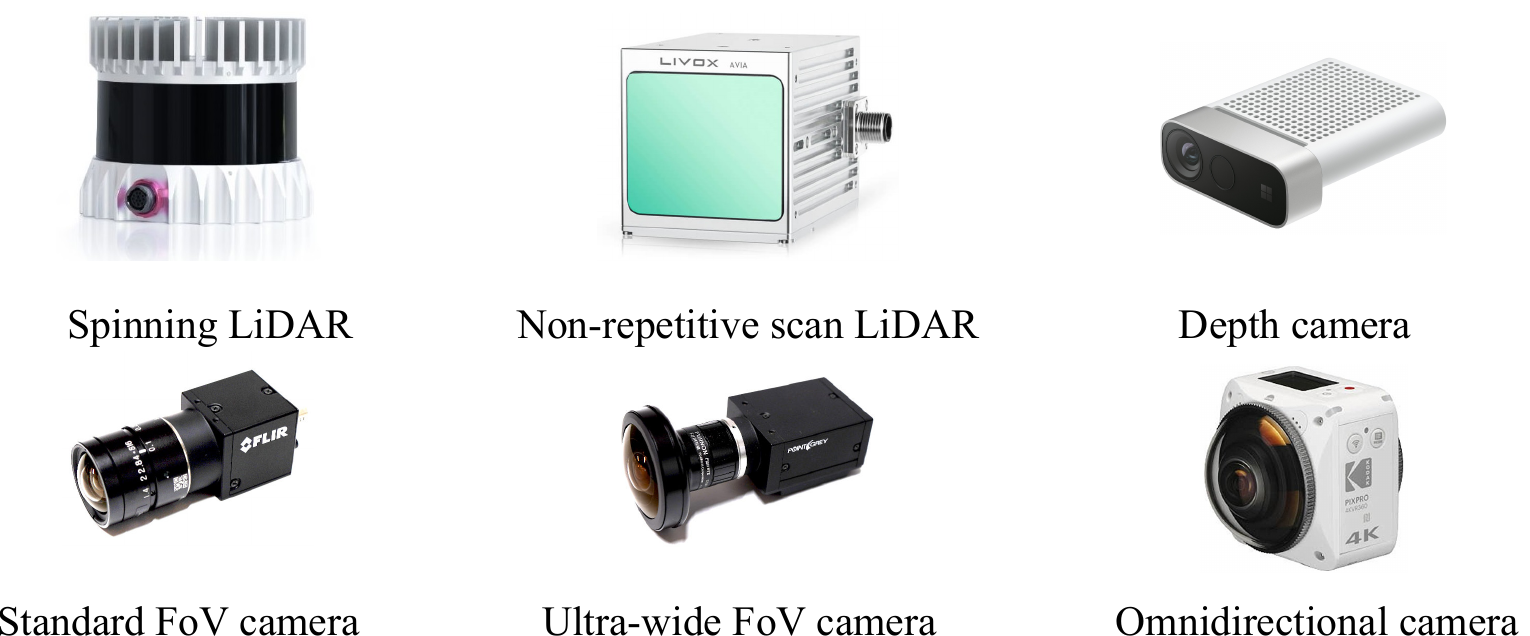}
  \subcaption{Supported LiDAR and camera models}
  \end{minipage}
  \vspace{1mm}
  \begin{minipage}[tb]{\linewidth}
  \centering
  \includegraphics[width=\linewidth]{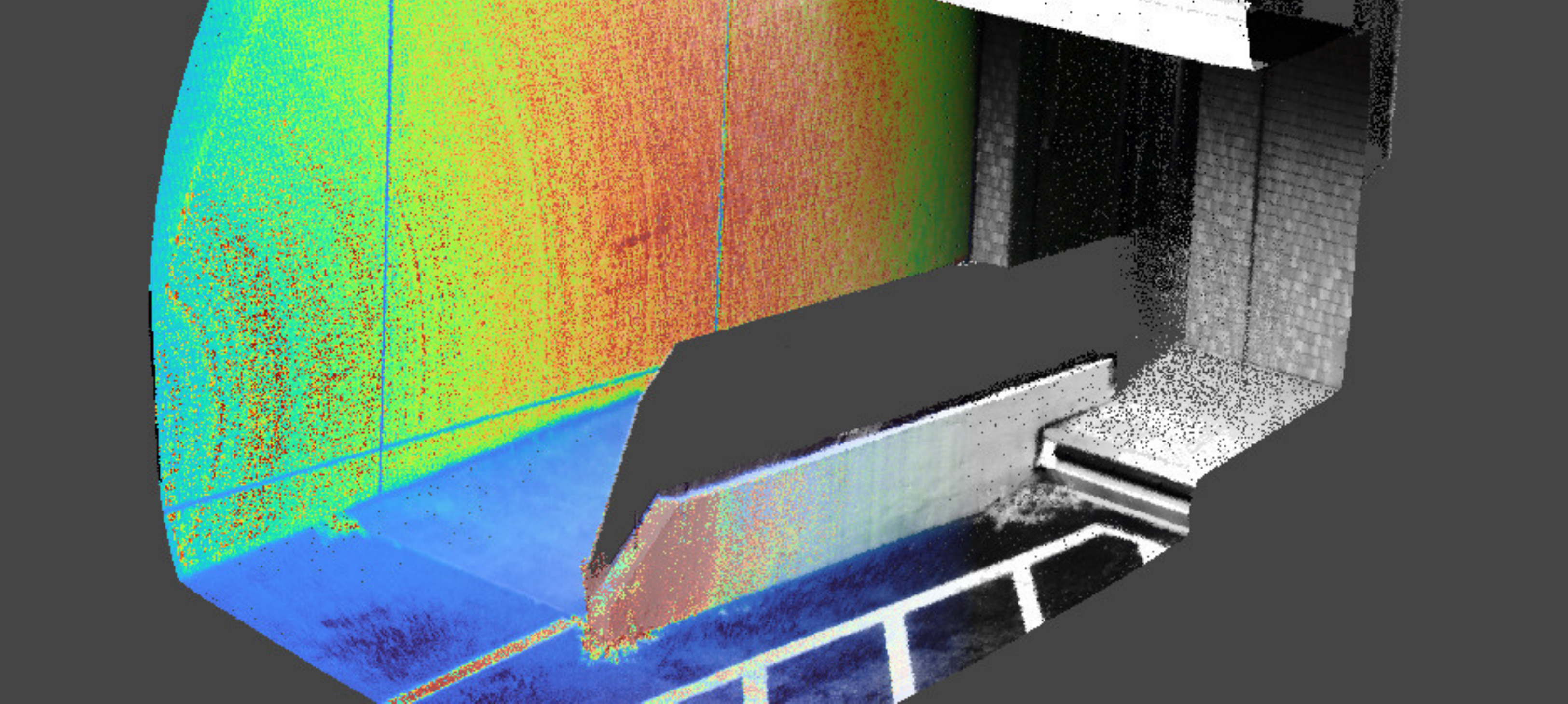}
  \subcaption{Calibration result (left to right: LiDAR to camera intensity)}
  \end{minipage}
  \caption{We present a complete LiDAR-camera calibration framework that can handle various LiDAR and camera models and calibrate the transformation between them from only a single pairing of a LiDAR point cloud and a camera image. The pixel-level direct alignment algorithm enables high-quality LiDAR-camera data fusion.}
  \label{fig:image}
\end{figure}

As a benefit to the robotics community, herein we present a new complete LiDAR-camera calibration toolbox that has the following features:
\begin{itemize}
  \item {\bf Generalizable:} The proposed toolbox is sensor-model-independent and can handle various LiDAR and camera projection models, including spinning and non-repetitive scan LiDARs, and pinhole, fisheye, and omnidirectional projection cameras, as shown in Fig. \ref{fig:image} (a).
  \item {\bf Target-less:} The proposed calibration algorithm does not require a calibration target but uses the environment structure and texture for calibration.
  \item {\bf Single-shot:} At a minimum, only one pairing of a LiDAR point cloud and a camera image is required for calibration. Optionally, multiple LiDAR-camera data pairs can be used to further improve the calibration accuracy.
  \item {\bf Automatic:} To make the calibration process fully automatic, the system includes an initial guess estimation algorithm with cross-modal 2D-3D correspondence matching based on SuperGlue \cite{Sarlin2020}.
  \item {\bf Accurate and robust:} A pixel-level direct LiDAR-camera registration algorithm is employed to robustly and accurately perform LiDAR-camera calibration in environments without rich geometrical features, where existing edge alignment-based algorithms \cite{Yuan2021} would fail, as long as there exists mutual information between LiDAR and camera data.
\end{itemize}
To our knowledge, there is no open implementation that has all the above features, and we believe the release of this toolbox will be beneficial to the robotics community.

The main contributions of this paper are as follows:
\begin{itemize}
  \item We present a robust initial guess estimation algorithm based 2D-3D correspondence estimation. To take advantage of the recent graph neural network-based image matching \cite{Sarlin2020}, we generate a LiDAR intensity image with a virtual camera and find correspondences between the LiDAR intensity image and the camera image. An estimate of the LiDAR-camera transformation is then given by RANSAC and reprojection error minimization.
  \item For robust and accurate calibration, we combined a direct LiDAR-camera fine registration algorithm based on the normalized information distance (NID), a mutual-information (MI)-based cross-modal distance metric, with a view-based hidden points removal algorithm that filters out points that are occluded and should not be visible from the viewpoint of the camera.
  \item The entire system was carefully designed to be general to LiDAR and camera projection models so that it can be applied to various sensor models. 
  \item We released the code of the developed method as open source to benefit the community \footnote{The code is available at \url{https://github.com/koide3/direct_visual_lidar_calibration}}.
\end{itemize}

\section{Related Work}

LiDAR-camera extrinsic calibration methods are categorized into three approaches: 1) target-based, 2) motion-based, and 3) scene-based.

\subsection{Target-based calibration}

As in the well-known camera intrinsic calibration process, the target-based approach is the most natural way for LiDAR-camera extrinsic calibration. Once we obtain the 3D coordinates of points on a calibration target and their corresponding 2D coordinates projected in the image, we can easily estimate the LiDAR-camera transformation by solving the perspective-n-point problem. The challenge here is that it is often difficult to design and create a calibration target that can robustly and accurately be detected by both the LiDAR and the camera. Several studies used a 3D structured calibration target that is not as easy to create as the well-known chessboard pattern \cite{Beltran2022,Fang2021}. Although several other works used a planar pattern that is easy to create, they required manual annotation of LiDAR data \cite{Zhang} and multiple data acquisitions, resulting in a large amount of effort \cite{Zhou2018}.

\subsection{Motion-based calibration}

As in the hand-eye calibration problem, we can estimate the transformation between two frames on a rigid body based on their motions \cite{Tsai1989}. This approach can easily handle heterogeneous sensors and does not need an overlap between sensors \cite{Ishikawa2018}. However, it requires careful time synchronization, which is not always possible when, for example, we use an affordable web camera. Furthermore, we need to estimate the per-sensor motion as accurately as possible for better calibration results.

\subsection{Scene-based calibration}

Scene-based methods estimate the LiDAR-camera transformation by considering the consistency between pairs of LiDAR point clouds and camera images. LiDAR points are projected in the image space, and then their consistency is measured with pixel values based on some metrics.

Similar to the visual odometry estimation problem, there are two major approaches for LiDAR-camera data consistency evaluation: indirect and direct.

The indirect approach first extracts feature points (e.g., edge points) from both the camera and LiDAR data, and computes the reprojection error between 2D-3D corresponding points \cite{Yuan2021,Zhang2021,Liu2022}. While it exhibits good convergence due to the discriminative features and robust correspondence creation, it requires the environment to be geometry-rich to extract sufficient feature points. Because a majority of texture information for surfaces in the environment is discarded in this approach, it is less accurate compared with the direct approach.

The latter approach compares the pixel and point intensities directly. Because LiDAR and cameras usually exhibit very different intensity distributions, simply taking the difference between them does not work in practice. To overcome the difference in modalities, MI-based metrics have been used for LiDAR-camera data comparison \cite{Pandey2021,Pandey2014}. Because they do not use the difference between LiDAR and image intensities but instead consider their co-occurrence, they can robustly measure the consistency between LiDAR and camera intensity values. Following \cite{stewart2016a,Jeong2019}, we used the NID metric derived from MI such that it satisfies the metric space axioms and becomes more robust than MI \cite{stewart2016a}.

\section{Methodology}

\subsection{Overview}

\begin{figure}[tb]
 \centering
 \includegraphics[width=0.8\linewidth]{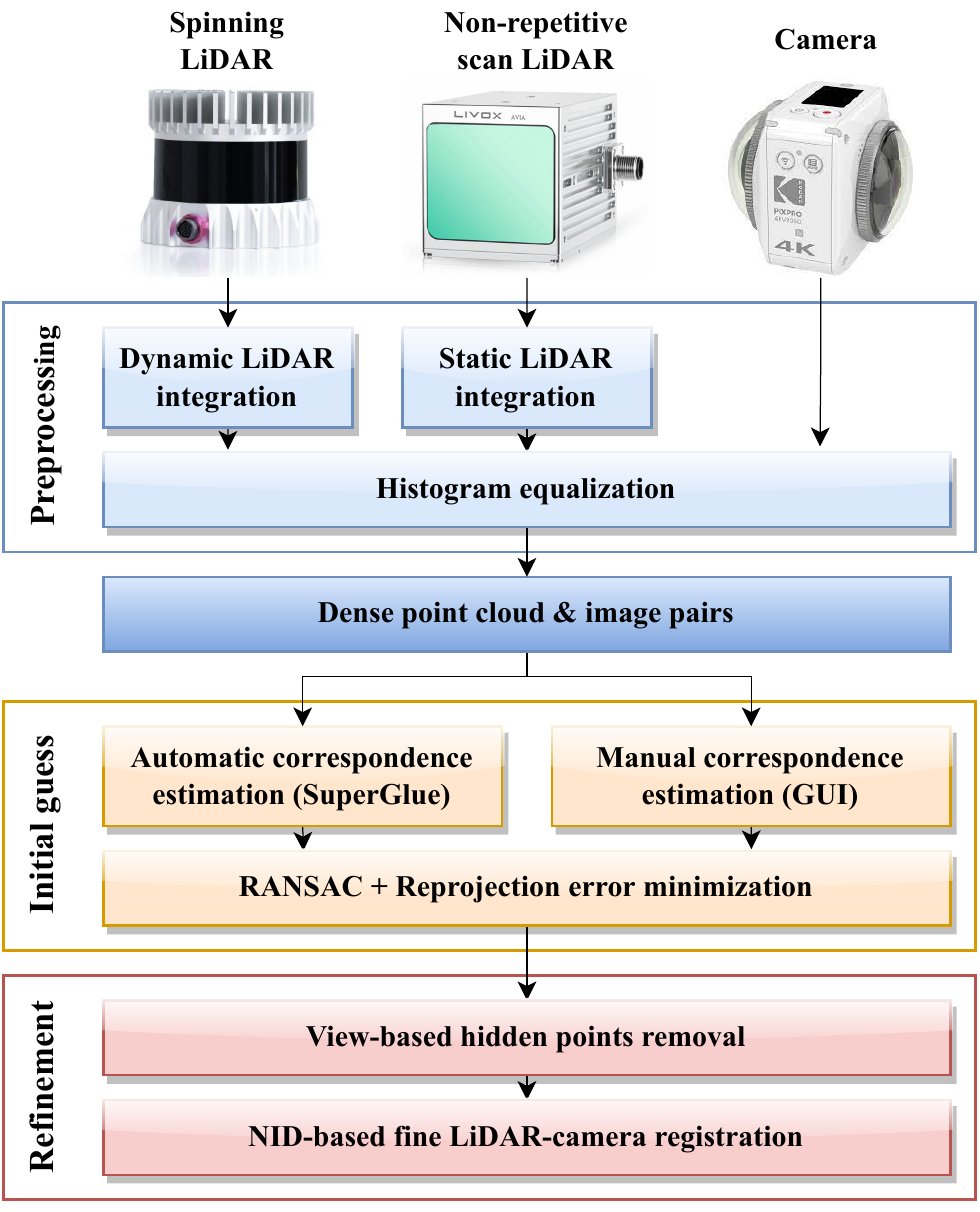}
 \caption{Overview of proposed LiDAR-camera calibration process. Input point clouds are merged to create dense point clouds using static and dynamic LiDAR point integrators. Given the densified point cloud and camera image, we find 2D-3D correspondences using the SuperGlue pipeline. We also provide an easy-to-use manual correspondence estimation tool. Given the 2D-3D correspondences, a rough estimate of the LiDAR-camera transformation is obtained via RANSAC and reprojection error minimization. Finally, we perform fine LiDAR-camera registration based on NID minimization.}
 \label{fig:system}
\end{figure} 
Fig. \ref{fig:system} shows an overview of the proposed LiDAR-camera calibration toolbox. 

To handle various LiDAR models with a unified processing pipeline, we first create a dense point cloud by merging multiple LiDAR frames. For non-repetitive scan LiDARs (e.g., Livox Avia), we simply accumulate points to densify the point cloud. For spinning LiDARs (e.g., Velodyne and Ouster LiDARs), we use a dynamic LiDAR point integration technique based on continuous-time ICP \cite{Dellenbach2022}.

From a paired densified point cloud and camera image, we then obtain a rough estimate of the LiDAR-camera transformation based on 2D-3D correspondences estimated with SuperGlue \cite{Sarlin2020}. We also provide an intuitive user interface for creating manual correspondence for fail-safe. Given the 2D-3D correspondences, we perform RANSAC and reprojection error minimization to obtain an initial estimate of the LiDAR-camera transformation.

Given the initial estimate of the LiDAR-camera transformation, we apply view-based hidden points removal to remove LiDAR points that should not be visible from the viewpoint of the camera. We then refine the LiDAR-camera transformation estimate via fine LiDAR-camera registration based on NID minimization.

\subsection{Notation}

Our goal is to estimate the transformation between LiDAR and camera coordinate frames ${^C{\bm T}_L}$ from pairings of LiDAR point clouds $\mathcal{P}_i = [{^L{\bm p}_1}, \cdots, {^L{\bm p}_N}]$ with point intensities $\mathcal{L}_i = [l_1, \cdots, l_N]$  and camera images $\mathcal{I}_i({\bm x}_j) = y_j$, where ${\bm x}_j \in \mathbb{R}^2$ are the pixel coordinates and $y_j$ is the pixel intensity. A point in the LiDAR frame ${^L{\bm p}_j}$ is transformed into the camera coordinate frame as ${^C{\bm p}_j} = {^C{\bm T}_L} {^L{\bm p}_j}$ and projected into the image space using a projection function $\pi$; ${\bm x}_j = \pi\left( {^C{\bm p}_j} \right)$. In this paper, we mainly use the conventional pinhole camera model with plumb-bob lens distortion and the omnidirectional equirectangular camera model \cite{Torii2009} as the projection function. Note that other major camera models, including ATAN \cite{Devernay2001}, fisheye \cite{Kannala2006}, and unified omnidirectional camera models \cite{Scaramuzza2006}, have been implemented and are supported in the developed toolbox.

\subsection{Preprocessing}

\begin{figure}[tb]
  \centering
  \begin{minipage}[tb]{\linewidth}
  \centering
  \includegraphics[width=0.9\linewidth]{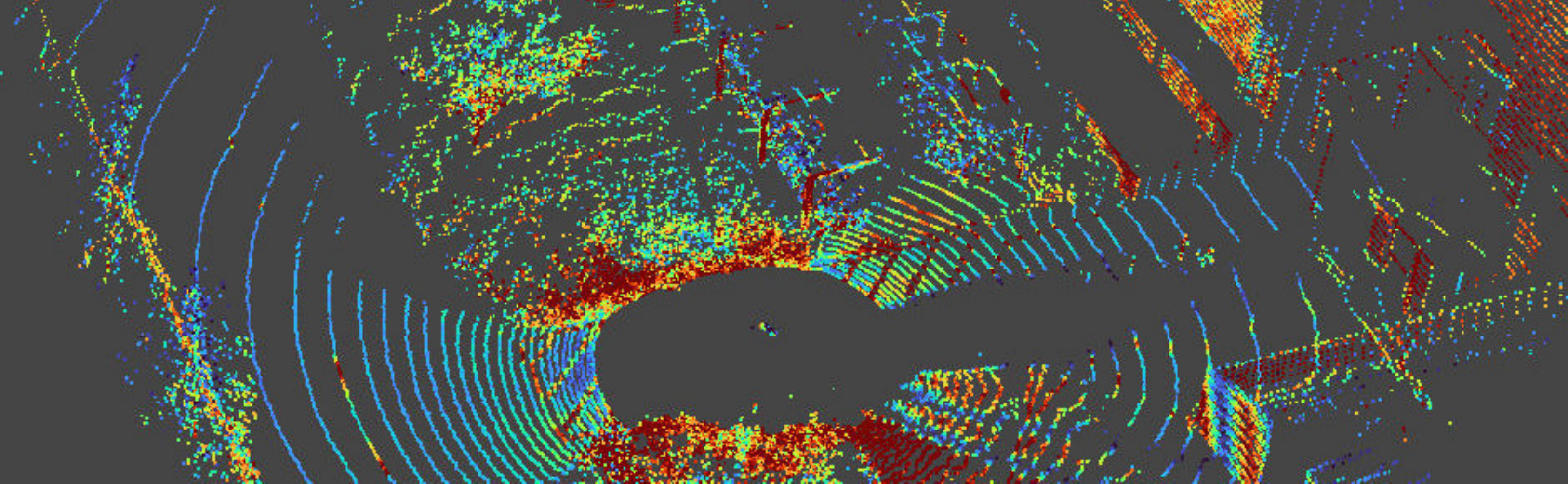}
  \subcaption{Single-scan point cloud}
  \end{minipage}
  \vspace{1mm}
  \begin{minipage}[tb]{0.9\linewidth}
  \centering
  \includegraphics[width=\linewidth]{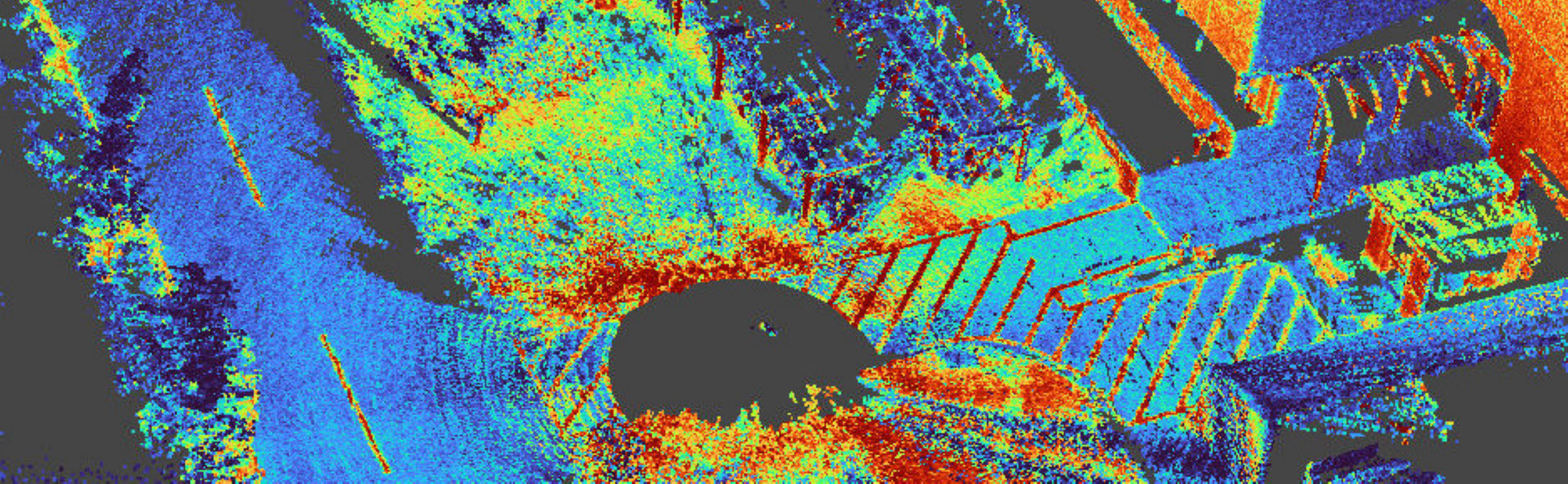}
  \subcaption{Densified point cloud}
  \end{minipage}
  \caption{Point cloud densification for spinning LiDARs. LiDAR integration enables the creation of a dense point cloud from a few seconds of dynamic LiDAR data. The densified point cloud exhibits rich geometrical and surface texture information.}
  \label{fig:ouster_points}
\end{figure}


Spinning LiDARs (e.g., Ouster OS1-64) exhibit a sparse and repetitive scan pattern, and it is difficult to extract meaningful geometrical and texture information from only a single scan (see Fig. \ref{fig:ouster_points} (a)). For such a LiDAR, we move the LiDAR in the up-down direction for a few seconds and accumulate points while compensating for the viewpoint change and point cloud distortion. To estimate the LiDAR motion, we use the CT-ICP algorithm, which jointly optimizes the LiDAR poses at the scan beginning and end by minimizing the distance between the current LiDAR scan and a model point cloud with the interpolated LiDAR pose. To efficiently create the target point cloud from past observations, we use the linear iVox \cite{Bai2022} structure, which simply keeps points in a linear container for each voxel. Based on the estimated LiDAR scan beginning and end poses, we correct the motion distortion on the input point cloud and create a dense point cloud by accumulating all points in the coordinate frame of the first scan. Fig. \ref{fig:ouster_points} (b) shows a densified point cloud after application of this dynamic LiDAR point integration process. We can see that the densified point cloud exhibits rich geometrical and texture information that was difficult to see in the single-scan point cloud.

For LiDARs with a non-repetitive scan mechanism (e.g., Livox Avia), we simply accumulate all scans into one frame, resulting in a dense point cloud, as shown in Fig. \ref{fig:image} (b). 

For the densified point cloud and the camera image, we apply histogram equalization because the NID metric used in the fine registration step works best with uniform intensity distributions.

\begin{figure}[tb]
  \centering
  \begin{minipage}[tb]{\linewidth}
  \centering
  \includegraphics[width=0.8\linewidth]{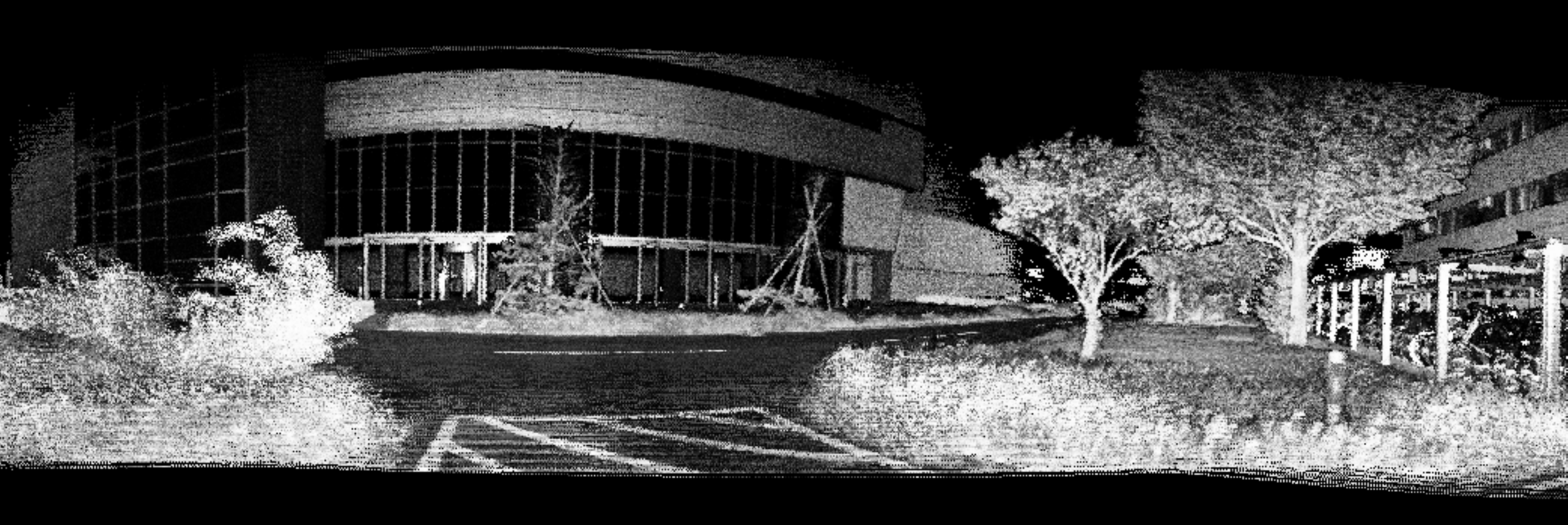}
  \subcaption{Ouster OS1-64 (estimated FoV: 178.6\textdegree)}
  \end{minipage}
  \vspace{1mm}
  \begin{minipage}[tb]{\linewidth}
  \centering
  \includegraphics[width=0.8\linewidth]{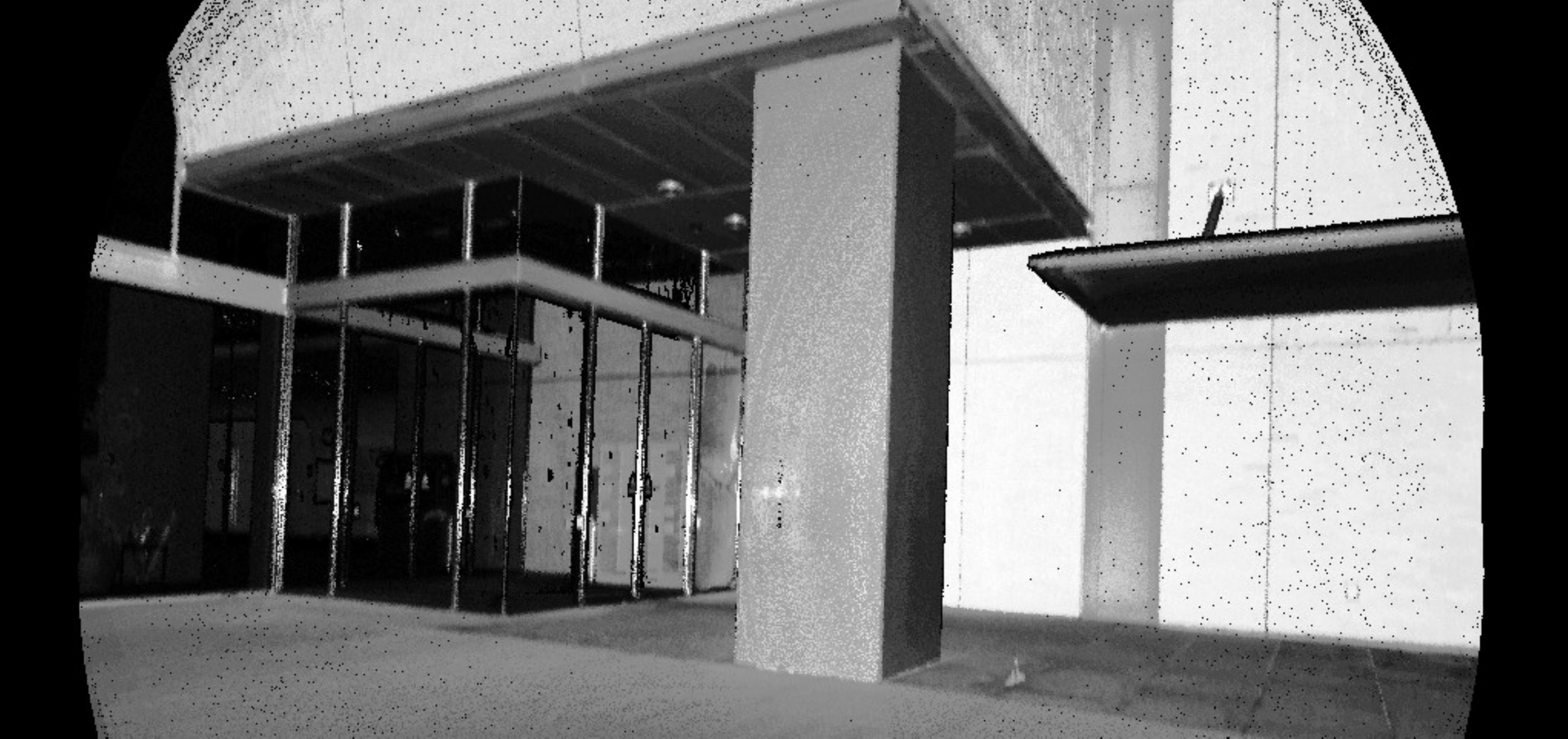}
  \subcaption{Livox Avia (estimated FoV: 76.2\textdegree)}
  \end{minipage}
  \caption{LiDAR intensity images rendered with virtual cameras (images are cropped due to space limitations). Either the pinhole or equirectangular projection model is selected depending on the FoV of the LiDAR.}
  \label{fig:lidar_images}
\end{figure}

\begin{figure}[tb]
  \centering
  \includegraphics[width=0.8\linewidth]{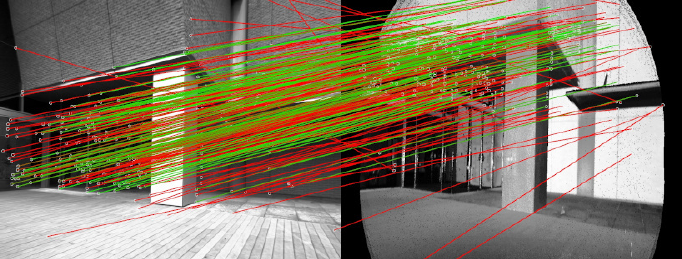}
  \caption{SuperGlue can find correspondences between LiDAR and camera images in different modalities with a very low sensitive matching threshold setting. The result, however, contains many false correspondences that need to be filtered out before pose estimation (green: inliers, red: outliers).}
  \label{fig:livox_matches}
\end{figure}

\subsection{Initial Guess Estimation}

To obtain a rough estimate of the LiDAR-camera transformation, we first obtain 2D-3D correspondences between the input images and point clouds and then estimate the LiDAR-camera transformation via RANSAC and reprojection error minimization.

To take advantage of the graph neural network-based image matching pipeline \cite{Sarlin2020}, we generate LiDAR intensity images from dense point clouds with a virtual camera model. To select the best projection model for rendering the entire point cloud, we first estimate the FoV of the LiDAR. We extract the convex hull of the input point cloud using the quickhull algorithm \cite{Barber1996} and then find the point pair with the maximum angle distance in the convex hull using a brute-force search. If the estimated FoV of the LiDAR is smaller than 150\textdegree, we create a virtual camera with the pinhole projection model. Otherwise, we create a virtual camera with the equirectangular projection model. With the virtual camera, we render the point cloud with intensity values to obtain LiDAR intensity images, as shown in Fig. \ref{fig:lidar_images}. Along with the intensity images, we also generate point index maps to efficiently look up pixel-wise 3D coordinates in the subsequent pose estimation step. Note that while we simply render each point without interpolation and gap filling, rendering results exhibit good appearance quality thanks to the densely accumulated point clouds.

To find correspondences between the LiDAR and camera intensity images, we use the SuperGlue pipeline \cite{Sarlin2020}. It first detects keypoints on images using SuperPoint \cite{DeTone2018} and then finds correspondences between the keypoints using a graph neural network. The weights pretrained on the MegaDepth dataset is used in this work. While SuperGlue can find correspondences between images in different modalities, we found that the matching threshold needs to be set to a very small value (e.g., 0.05) to obtain a sufficient number of correspondences. However, with this setting, we observed that many false correspondences are created, as shown in Fig. \ref{fig:livox_matches}.

\begin{algorithm}[tb]
\caption{Rotation-only RANSAC}
\label{alg:ransac}
\begin{algorithmic}[1]
\Input 2D keypoints $\mathcal{K} = [{\bm x}^K_1, \cdots, {\bm x}^K_{M}]$ and corresponding 3D coordinates $\mathcal{D} = [{^L {\bm p}_1^K}, \cdots, {^L {\bm p}_M^K}]$

\Function{EstimateRotationRANSAC}{$\mathcal{K}, \mathcal{D}$}
  \State ${^C{\bm d}_j} \gets \frac{ \pi^{-1} \left( {\bm x}_j^K \right) } { \| \pi^{-1} \left( {\bm x}_j^K \right) \| } $ \Comment{Bearing vectors of ${{\bm x}_j^K}$}
  \State ${^L{\bm d}_j} \gets \frac{ ^L{\bm p}_j^K } { \| ^L{\bm p}_j^K \| } $ \Comment{Bearing vectors of ${^L{\bm p}_j^K}$}

  \For{ $i \in [1, \cdots, N^{\text{iteration}}]$ }
    \State Randomly sample two correspondences $j_0$ and $j_1$
    \State ${^C{\bm R}_{L}} \gets $ \Call{FindRotationLSQ}{$j_0$, $j_1$}
    \State $N \gets $ count of ${\bm x}_j^i$ s.t. $|\pi\left( {^C{\bm R}_L} {^L{\bm p}_j^K} \right) - {\bm x}_j^K| < \alpha $
    \If{$i = 1$ or $N > N^*$}
      \State $N^* \gets N$
      \State ${^C{\bm R}_L^*} \gets {^C{\bm R}_L}$
    \EndIf
  \EndFor

  \State \textbf{return} ${^C{\bm R}_L^*}$
\EndFunction

\Function{EstimateRotationLSQ}{$j_0$, $j_1$} \cite{Umeyama1991}
  \State ${\bm A} \gets [ {^C{\bm d}_{j_0}} {^C{\bm d}_{j_1}} ]$, ${\bm B} \gets [ {^L{\bm d}_{j_0}} {^L{\bm d}_{j_1}} ]$
  \State ${\bm U} {\bm \Sigma} {\bm V}^* = {\bm A}{\bm B}$ \Comment{SVD}
  \State $s \gets 1$ \textbf{if} $|{\bm U}| |{\bm V}| >= 0$ \textbf{else} $-1$
  \State \textbf{return} ${\bm U} \text{diag}([1, 1, s]) {\bm V}^*$
\EndFunction

\end{algorithmic}
\end{algorithm}

Given 2D keypoint coordinates $\mathcal{K} = [{\bm x}^K_1, \cdots, {\bm x}^K_{M}]$ and corresponding 3D coordinates in the LiDAR frame $\mathcal{D} = [{^L {\bm p}_1^K}, \cdots, {^L {\bm p}_M^K}]$, we first perform the rotation-only RANSAC described in Alg. \ref{alg:ransac} to robustly deal with outlier correspondences. Given the estimated rotation as an initial guess, we then obtain the 6 DoF LiDAR-camera transformation $^C\tilde{{\bm T}}_L$ by minimizing the reprojection errors of all keypoints using the Levenberg-Marquardt optimizer:

\begin{align}
{^C\tilde{{\bm T}}_L} &= \argmin_{^C{\bm T}_L} \sum_{j=1}^{M} \rho \left( \| \pi\left( {^C{\bm T}_L} {^L{\bm p}_j^K} \right) - {\bm x}_j^K \|^2 \right),
\end{align}
where $\rho$ is the Cauchy robust kernel.

\subsection{NID-based Direct LiDAR-Camera Registration}

Some points in the LiDAR point cloud can be occluded and not visible from the camera due to the viewpoint difference. If we simply project all the LiDAR points, such points can cause false correspondences and affect the calibration result, as discussed in \cite{Yuan2021}. To avoid this problem, we apply efficient view-based hidden point removal to filter out LiDAR points that should not be visible from the viewpoint of the camera. With the current estimate of the LiDAR-camera transformation, we project the LiDAR points in the image and keep only the point with the minimum distance for each pixel (i.e., depth buffer testing). From the projected image that retains only points visible from the camera, we obtain a 3D point cloud and use it for fine registration.

We then perform direct LiDAR-camera registration based on the NID metric \cite{stewart2016a}. To compute the NID, we transform LiDAR points ${^L{\bm p}_j} \in \mathcal{P}_i$ in the camera frame and project them into the image space ${\bm x}_j = \pi\left( {^C{\bm T}_L} {^L{\bm p}_j} \right)$. From the LiDAR point intensities $l_j$ and corresponding pixel intensities $\mathcal{I}_i\left( {\bm x}_j \right)$, we create $\text{P}(\mathcal{L}_i)$, $\text{P}(\mathcal{I}_i)$, and $\text{P}(\mathcal{L}_i, \mathcal{I}_i)$, which are marginal and joint histograms of LiDAR and pixel intensities, and calculate their entropies $\text{H}(\mathcal{L}_i)$, $\text{H}(\mathcal{I}_i)$, and $\text{H}(\mathcal{L}_i, \mathcal{I}_i)$ as follows:

\begin{align}
\text{H}\left( X \right) = \sum_{x \in X} p(x) \log p(x),
\end{align}
where $x$ is each bin in the histogram. The NID between $\mathcal{L}_i$ and $\mathcal{I}_i$ is then defined as follows:
\begin{align}
\label{eq:nid}
\text{NID}\left(\mathcal{L}_i, \mathcal{I}_i \right) &= \frac{ \text{H}\left( \mathcal{L}_i, \mathcal{I}_i \right) - \text{MI}\left(\mathcal{L}_i; \mathcal{I}_i \right) } { \text{H}\left( \mathcal{L}_i, \mathcal{I}_i \right)}, \\
\text{MI}\left(\mathcal{L}_i; \mathcal{I}_i \right) &= \text{H}\left( \mathcal{L}_i \right) + \text{H}\left( \mathcal{I}_i \right) -  \text{H}\left( \mathcal{L}_i, \mathcal{I}_i \right),
\end{align}
where $\text{MI}\left( \mathcal{L}_i; \mathcal{I}_i \right)$ is the mutual information between $\mathcal{L}_i$ and $\mathcal{I}_i$. By using the Nelder-Mead optimizer, we find the LiDAR-camera transformation that minimizes Eq. \ref{eq:nid}.

Because LiDAR points visible from the camera can change as the LiDAR-camera transformation is updated, we iterate the view-based hidden points removal and NID-based LiDAR-camera registration until the displacement of the transformation update converges.

\section{Experiment}

\begin{figure}[tb]
  \centering
  \begin{minipage}[tb]{0.48\linewidth}
  \centering
  \includegraphics[height=2.8cm]{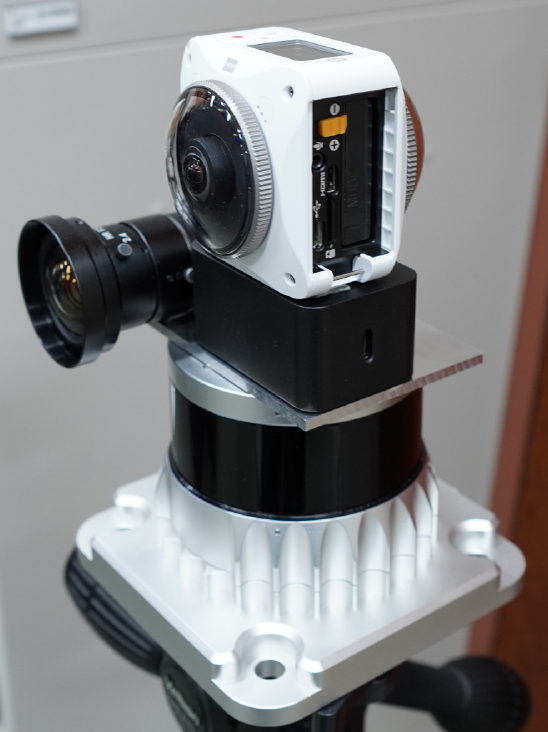}
  \subcaption{Ouster OS1-64}
  \end{minipage}
  \begin{minipage}[tb]{0.48\linewidth}
  \centering
  \includegraphics[height=2.8cm]{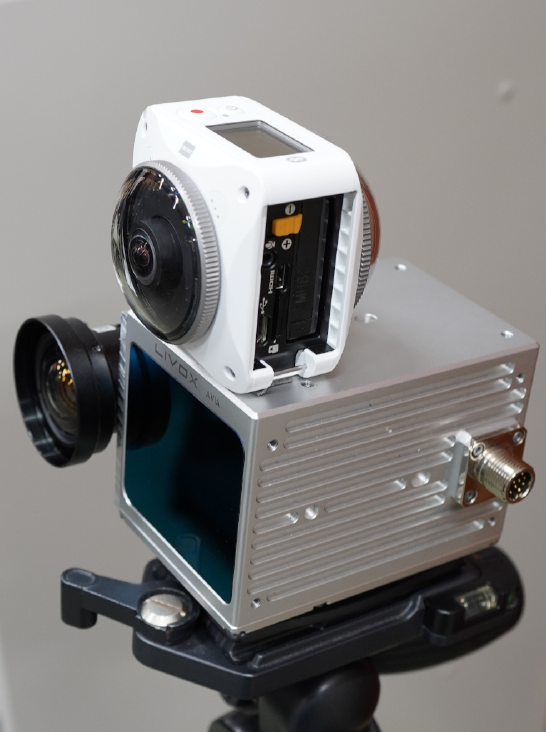}
  \subcaption{Livox Avia}
  \end{minipage}
  \caption{Sensor configurations for LiDAR-camera calibration experiments.}
  \label{fig:lidar_cams}
\end{figure}

\begin{figure}[tb]
  \centering
  \includegraphics[height=2.8cm]{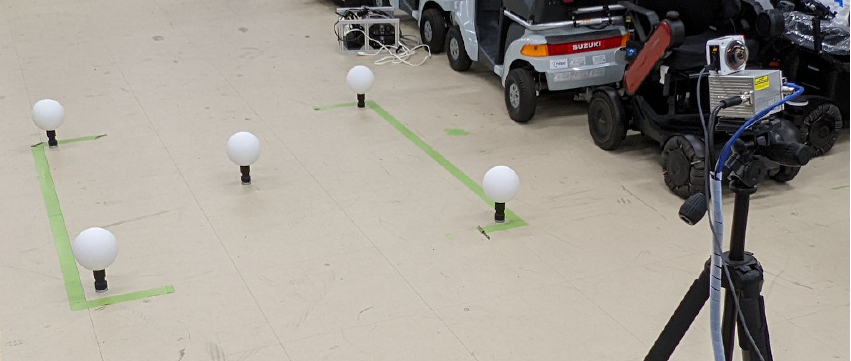}
  \caption{Reference LiDAR-camera transformations were measured using high-reflectivity sphere targets; 2D and 3D positions of the targets were manually annotated and the transformation was estimated by minimizing their reprojection errors.}
  \label{fig:reference}
\end{figure}

We evaluated the proposed calibration toolbox with all four combinations of the spinning and non-repetitive scan LiDARs (Ouster OS1-64 and Livox Avia) and pinhole and omnidirectional cameras (Omron Sentech STC-MBS202POE and Kodak PixPro 4KVR360) shown in Fig. \ref{fig:lidar_cams}. For each combination, we recorded 15 pairs of LiDAR point clouds and camera images in indoor and outdoor environments, and we ran the proposed calibration process for each pair (i.e., single-shot calibration).

As a reference, we estimated the LiDAR-camera transformation using survey-grade high-reflectivity sphere targets, as shown in Fig.\ref{fig:reference}. We manually annotated the 2D and 3D positions of the targets to find the LiDAR-camera transformations that minimized the target reprojection errors. From visual inspection, we confirmed that the estimated transformations closely describe the projection of the cameras. We used the estimated transformations as the ``pseudo'' ground truth.

\begin{table}[tb]
  \caption{Calibration errors for Ouster OS1-64 and pinhole camera}
  \label{tab:ouster_pinhole}
  \centering
  \scriptsize
  \begin{tabular}{g|ggg|gg}
  \toprule
  \rowcolor{white}                        & \multicolumn{3}{c|}{Proposed}               & \multicolumn{2}{c}{Edge-based \cite{Yuan2021}} \\
  \rowcolor{white} Data & Init. guess & Trans. [m] & Rot. [\textdegree]  & Trans. [m] & Rot. [\textdegree]  \\
  \midrule
  \rowcolor{white} 00   & \checkmark & 0.019 & 0.688 & 0.043 & 0.135 \\
                   01   & \xmark     & 0.028 & 0.348 & 0.029 & 0.685 \\
  \rowcolor{white} 02   & \xmark     & 0.014 & 0.180 & 0.035 & 0.490 \\
                   03   & \checkmark & 0.006 & 0.663 & 0.056 & 0.377 \\
  \rowcolor{white} 04   & \xmark     & 0.029 & 0.400 & 0.158 & 1.499 \\
                   05   & \checkmark & 0.023 & 0.450 & 0.170 & 2.334 \\
  \rowcolor{white} 06   & \checkmark & 0.033 & 0.613 & 0.136 & 1.034 \\
                   07   & \checkmark & 0.017 & 0.392 & 0.033 & 0.154 \\
  \rowcolor{white} 08   & \checkmark & 0.048 & 0.386 & 0.520 & 1.458 \\
                   09   & \checkmark & 0.007 & 0.251 & 0.317 & 1.254 \\
  \rowcolor{white} 10   & \checkmark & 0.064 & 0.210 & 0.191 & 1.463 \\
                   11   & \checkmark & 0.011 & 0.146 & 0.384 & 0.768 \\
  \rowcolor{white} 12   & \checkmark & 0.325 & 0.530 & 0.226 & 0.524 \\
                   13   & \checkmark & 0.010 & 0.137 & 1.211 & 7.344 \\
  \rowcolor{white} 14   & \checkmark & 0.015 & 0.217 & 0.034 & 0.421 \\
  \midrule
  \rowcolor{white} Avg. & 12 / 15    & 0.043 & 0.374 & 0.236 & 1.329 \\
  \bottomrule
  \end{tabular}
  \\ \vspace{1mm}
  \checkmark and \xmark respectively represent the success and failure of the initial guess.
\end{table}

\begin{table}[tb]
  \caption{Calibration errors for Livox Avia and pinhole camera}
  \label{tab:livox_pinhole}
  \centering
  \scriptsize
  \begin{tabular}{g|ggg|gg}
  \toprule
  
  \rowcolor{white}      & \multicolumn{3}{c|}{Proposed}               & \multicolumn{2}{c}{Edge-based \cite{Yuan2021}} \\
  \rowcolor{white} Data & Init. guess & Trans. [m] & Rot. [\textdegree]  & Trans. [m] & Rot. [\textdegree]  \\
  \midrule
  \rowcolor{white} 00   & \checkmark & 0.047  & 0.478 & 1.054  & 6.964 \\
                   01   & \checkmark & 0.162  & 0.885 & 0.152  & 1.336 \\
  \rowcolor{white} 02   & \checkmark & 0.028  & 0.356 & 1.587  & 14.278 \\
                   03   & \checkmark & 0.098  & 0.757 & 0.200  & 3.480 \\
  \rowcolor{white} 04   & \checkmark & 0.124  & 1.430 & 0.065  & 0.933 \\
                   05   & \checkmark & 0.027  & 0.466 & 0.105  & 3.852 \\
  \rowcolor{white} 06   & \checkmark & 0.032  & 0.410 & 0.128  & 1.577 \\
                   07   & \checkmark & 0.026  & 0.273 & 0.216  & 3.609 \\
  \rowcolor{white} 08   & \checkmark & 0.031  & 0.270 & 0.358  & 4.450 \\
                   09   & \checkmark & 0.054  & 0.665 & 0.117  & 1.522 \\
  \rowcolor{white} 10   & \checkmark & 0.071  & 0.887 & 0.214  & 1.590 \\
                   11   & \checkmark & 0.029  & 0.412 & 0.085  & 0.651 \\
  \rowcolor{white} 12   & \checkmark & 0.046  & 0.297 & 0.181  & 2.133 \\
                   13   & \checkmark & 0.080  & 0.645 & 0.170  & 2.152 \\
  \rowcolor{white} 14   & \checkmark & 0.032  & 0.452 & 0.210  & 3.934 \\
  \midrule
  \rowcolor{white} Avg. & 15 / 15    & 0.059  & 0.579 & 0.323  & 3.497 \\
  \bottomrule
  \end{tabular}
  \\ \vspace{1mm}
  \checkmark and \xmark respectively represent the success and failure of the initial guess.
\end{table}

\begin{table}[tb]
  \caption{Calibration errors for Ouster OS1-64, Livox Avia, and omnidirectional camera}
  \label{tab:pixpro}
  \centering
  \scriptsize
  \begin{tabular}{g|ggg|ggg}
  \toprule
  
  \rowcolor{white}      & \multicolumn{3}{c|}{Ouster OS1-64 \& Omnidirectional}    & \multicolumn{3}{c}{Livox Avia \& Omnidirectional} \\
  \rowcolor{white} Data & Init. & Trans. [m] & Rot. [\textdegree]  & Init. & Trans. [m] & Rot. [\textdegree]  \\
  \midrule
  \rowcolor{white} 00   & \checkmark & 0.123 & 1.018 & \checkmark & 0.029 & 0.515 \\
                   01   & \checkmark & 0.111 & 0.355 & \xmark     & 0.059 & 0.902 \\
  \rowcolor{white} 02   & \checkmark & 0.085 & 0.747 & \checkmark & 0.071 & 0.518 \\
                   03   & \checkmark & 0.045 & 0.744 & \checkmark & 0.037 & 0.958 \\
  \rowcolor{white} 04   & \checkmark & 0.061 & 0.664 & \xmark     & 0.067 & 1.265 \\
                   05   & \checkmark & 0.033 & 0.632 & \xmark     & 0.113 & 1.447 \\
  \rowcolor{white} 06   & \checkmark & 0.057 & 0.540 & \xmark     & 0.070 & 0.312 \\
                   07   & \checkmark & 0.038 & 0.410 & \checkmark & 0.056 & 1.466 \\
  \rowcolor{white} 08   & \checkmark & 0.035 & 0.356 & \checkmark & 0.079 & 0.635 \\
                   09   & \checkmark & 0.087 & 0.349 & \checkmark & 0.407 & 1.164 \\
  \rowcolor{white} 10   & \checkmark & 0.087 & 0.428 & \xmark     & 0.161 & 0.753 \\
                   11   & \checkmark & 0.062 & 0.769 & \checkmark & 0.098 & 0.681 \\
  \rowcolor{white} 12   & \checkmark & 0.079 & 0.497 & \checkmark & 0.212 & 0.936 \\
                   13   & \checkmark & 0.095 & 2.939 & \checkmark & 0.033 & 0.409 \\
  \rowcolor{white} 14   & \checkmark & 0.044 & 0.415 & \checkmark & 0.028 & 0.149 \\
  \midrule
  \rowcolor{white} Avg. & 15 / 15    & 0.069 & 0.724 & 10 / 15    & 0.101 & 0.807 \\
  \bottomrule
  \end{tabular}
  \\ \vspace{1mm}
  \checkmark and \xmark respectively represent the success and failure of the initial guess.
\end{table}





\begin{table}[tb]
  \caption{Multi-data calibration errors}
  \label{tab:multi}
  \centering
  \scriptsize
  \begin{tabular}{lcccc}
  \toprule
          & \multicolumn{2}{c}{Pinhole} & \multicolumn{2}{c}{Omnidirectional}  \\
          & Trans [m]  & Rot [\textdegree]   & Trans [m]   & Rot [\textdegree] \\
  \midrule
  Ouster  & 0.034      & 0.414               & 0.082       & 0.425 \\
  Livox   & 0.010      & 0.132               & 0.011       & 0.400 \\
  \bottomrule
  \end{tabular}
\end{table}

\begin{table}[tb]
  \caption{Processing time}
  \label{tab:proctime}
  \centering
  \scriptsize
  \begin{tabular}{gg|gggg}
  \toprule
  \rowcolor{white} LiDAR & Camera           & Preprocessing & Init. guess & Calibration & Total \\
  \midrule
  \rowcolor{white} Ouster & Pinhole   & 56.6 s & 10.3 s & 28.8 s  & 95.7 s \\
                   Ouster & Omnidir.  & 56.3 s & 11.2 s & 181.5 s & 249.0 s \\
  \rowcolor{white} Livox  & Pinhole   & 9.3 s  & 10.6 s & 54.5 s  & 74.4 s \\
                   Livox  & Omnidir.  & 9.2 s  & 9.7 s  & 84.3 s  & 103.2 s \\
  \bottomrule
  \end{tabular}
\end{table}

\begin{figure}[tb]
  \centering
  \includegraphics[width=\linewidth]{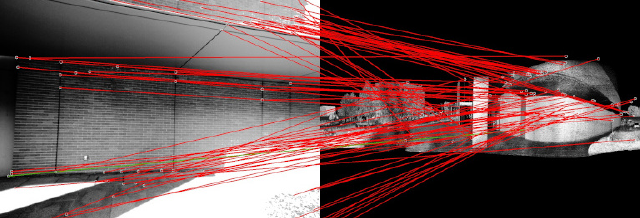}
  \caption{Failure case for initial guess estimation (Ouster 02). (green: inliers, red: outliers)}
  \label{fig:failed_init_guess}
\end{figure}

Table \ref{tab:ouster_pinhole} summarizes the calibration results for the combination of the Ouster LiDAR and pinhole camera. For the initial guess estimation, if the translation and rotation errors are smaller than 0.5 m and 1.0\textdegree, respectively, we consider the initial guess estimation as successful. As shown in Table \ref{tab:ouster_pinhole}, the proposed algorithm provided a reasonable initial guess for 12 out of 15 data correlations. Fig. \ref{fig:failed_init_guess} shows the correspondences of a failure case (Dataset 02). We can see that the estimated correspondences contain many false correspondences due to the flat and repetitive environment structure. Note that this kind of data pairing, where correct correspondences are difficult to find, can automatically be filtered out by RANSAC in multiple-data calibration and does not affect the estimation result. For data that resulted in a failed initial guess, we manually annotated 2D-3D correspondences and re-estimated the transformation for evaluation of the fine registration algorithm.

The proposed fine registration algorithm worked well for all the data and achieved 0.043 m and 0.374\textdegree \ calibration errors on average. As a baseline, we also applied the state-of-the-art edge alignment-based calibration method \cite{Yuan2021}. For this method, we used the reference LiDAR-camera transformation as the initial transformation, and thus it was evaluated with an almost ideal initial guess. However, as can be seen in Table \ref{tab:ouster_pinhole}, it showed large calibration errors for several data correlations, resulting in worse average calibration errors (0.236 m and 1.329\textdegree). This was because both image and point cloud edge extraction in \cite{Yuan2021} were very sensitive to environment changes and could not properly extract edge points in several environments.

As shown in Table \ref{tab:livox_pinhole}, for the combination of the Livox LiDAR and pinhole camera, the initial guess estimation successfully provided good initial LiDAR-camera transformations for all data pairings, and the fine registration algorithm achieved average calibration errors of 0.069 m and 0.724\textdegree, which were better than those for the edge-based calibration \cite{Yuan2021} (0.323 m and 3.497\textdegree).

Table \ref{tab:pixpro} summarizes the calibration results for the omnidirectional camera. While it showed a good initial guess success rate (15 / 15) and low calibration errors for the Ouster LiDAR (0.069 m and 0.724\textdegree), both the initial guess estimation and the fine LiDAR-camera registration tended be degraded for the combination of the Livox LiDAR and omnidirectional camera (10 / 15 success rate, 0.101 m and 0.807\textdegree). This is because only a small portion of the omnidirectional camera images was used for calibration due to the very different FoVs of the LiDAR and camera, and the resolution of the images was not sufficient to represent fine environmental details with this limited FoV. We think that the calibration accuracy can be improved by using higher-resolution or multiple images.

Finally, we performed multi-data calibration with all data pairings and evaluated the calibration errors for all the LiDAR-camera combinations. Table \ref{tab:multi} summarizes the calibration errors. We can see that good calibration results were obtained even for the combination of the Livox and omnidirectional camera (0.011 m and 0.400\textdegree). For the combination of the Ouster LiDAR and pinhole camera, we observed the best calibration accuracy (0.010 m and 0.132\textdegree). This is because in this combination, the LiDAR and camera have similar FoVs and most of input data were used for calibration while some points and image regions are not used in other combinations due to the difference of FoV.

 Table \ref{tab:proctime} shows the processing times for each calibration step. Depending on the combination of the LiDAR and camera models, it took from 74 to 249 s to calibrate the LiDAR-camera transformation from 15 data pairings. Although the combination of the Ouster LiDAR and omnidirectional camera took a longer time because both the sensors have 360\textdegree \ FoVs and need projection of most of the points, we consider it is in a reasonable level for offline calibration.

\section{Conclusion}

We developed a general LiDAR-camera calibration toolbox. For a fully automatic calibration process, we used image matching-based initial guess estimation. The initial estimate was then refined by a NID-based direct LiDAR-camera registration algorithm. The experimental results showed that the toolbox can accurately calibrate the transformation between spinning and non-repetitive scan LiDARs and pinhole and omnidirectional cameras.

\balance

\bibliographystyle{IEEEtran}
\bibliography{icra2023}

\end{document}